\documentclass[letterpaper, 10 pt, conference]{ieeeconf}  

\IEEEoverridecommandlockouts                              

\usepackage{cite}
\usepackage{amsmath,amssymb,amsfonts}
\usepackage{algorithmic}
\usepackage{graphicx}
\usepackage{subfig}
\usepackage{url}

\usepackage{textcomp}
\usepackage{xcolor}

\usepackage{float}
\usepackage{booktabs}
\usepackage{caption}
\usepackage{threeparttable}
\usepackage{adjustbox}
\title{\LARGE \bf
Vision-based Wearable Steering Assistance for People with Impaired Vision in Jogging
}

\author{Xiaotong Liu, Binglu Wang,~\IEEEmembership{Member,~IEEE,} Zhijun Li,~\IEEEmembership{Fellow,~IEEE}
\thanks{This work was supported in part by the National Natural Science Foundation of China under Grant 62133013, U22A2060, 62333007 and 62303151; in part by the Project funded by China Postdoctoral Science Foundation 2023M733385; in part by Anhui Provincial Natural Science Foundation 2308085QF206. (Corresponding author: Zhijun Li)}
\thanks{Xiaotong Liu is with the Institute of Advanced Technology, University
	of Science and Technology of China, Hefei 230026, China. liuxiaotong@mail.ustc.edu.cn}%
\thanks{Binglu Wang is with the School of Information and Electronic, Beijing Institute of Technology, Beijing 100081, China. wbl921129@gmail.com}%
\thanks{Zhijun Li is with School of Mechanical Engineering, Tongji University, Shanghai 200092, China. zjli@ieee.org}
}

\begin{document}

\maketitle
\thispagestyle{empty}
\pagestyle{empty}

\begin{abstract}
	
Outdoor sports pose a challenge for people with impaired vision. 
The demand for higher-speed mobility inspired us to develop a vision-based wearable steering assistance. 
To ensure broad applicability, we focused on a representative sports environment, the athletics track.
Our efforts centered on improving the speed and accuracy of perception, enhancing planning adaptability for the real world, and providing swift and safe assistance for people with impaired vision. In perception, we engineered a lightweight multitask network capable of simultaneously detecting track lines and obstacles. 
Additionally, due to the limitations of existing datasets for supporting multi-task detection in athletics tracks, we diligently collected and annotated a new dataset (MAT) containing 1000 images.
In planning, we integrated the methods of sampling and spline curves, addressing the planning challenges of curves. Meanwhile, we utilized the positions of the track lines and obstacles as constraints to guide people with impaired vision safely along the current track.
Our system is deployed on an embedded device, Jetson Orin NX. Through outdoor experiments, it demonstrated adaptability in different sports scenarios, assisting users in achieving free movement of 400-meter at an average speed of 1.34 m/s, meeting the level of normal people in jogging.
Our MAT dataset is publicly available from
\url{https://github.com/snoopy-l/MAT}

\end{abstract}


\section{INTRODUCTION}

According to statistics from the World Health Organization (WHO), nearly 253 million people worldwide are facing visual impairments \cite{11}, which affects both their physical health and quality of life. 
Sports, like jogging, becomes particularly challenging for people with impaired vision, as they lack the visual cues essential for safe navigation. Even in Paralympic athletics, visually impaired athletes rely on guide runners, rendering independent running unattainable.

Advancements in artificial intelligence and wearable devices led to the development of assisting the blind and visually impaired (BVI) in overcoming common challenges in their daily lives\cite{14}. For instance, \cite{12} provides navigation speeds of 0.21 m/s, \cite{sr} facilitated walking speeds of 0.6 m/s, and \cite{sp} enables people with impaired vision to walk through hallways at speeds of 1 m/s.
These studies have contributed to enhancing the mobility of them. 
Nonetheless, there remains a dearth of exploration in sports and exercise, leaving the quest for faster independent mobility largely uncharted. Facilitating assistance for people with impaired vision in sports can also be beneficial for their physical health. Consequently, the design of wearable steering assistance tailored for people with impaired vision in sports holds crucial practical significance.

\begin{figure}[t]
	\centerline{\includegraphics[width=1.87in]{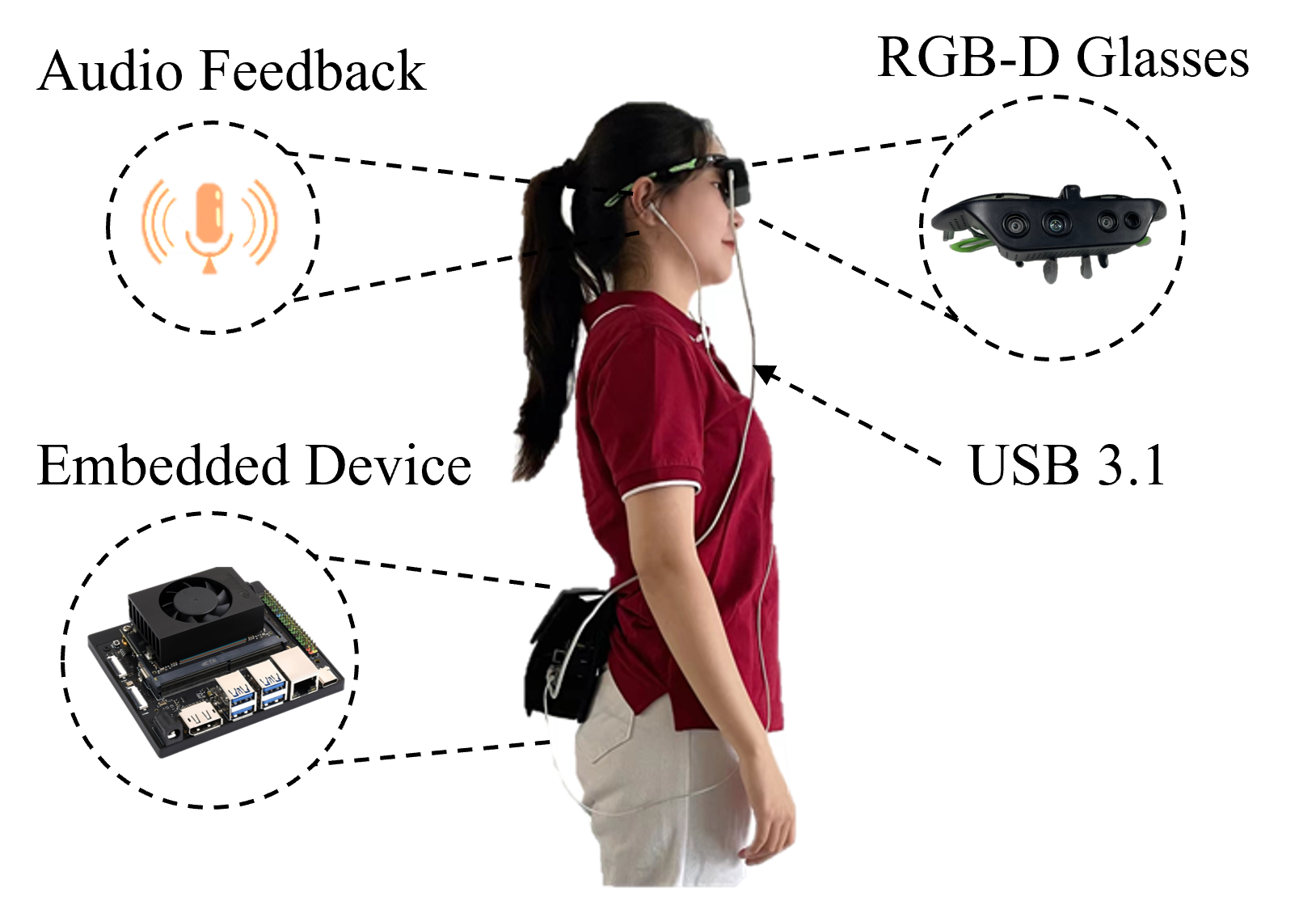}}
	\caption{The hardware of wearable steering assistance. The assistance consists of a pair of RGB-D glasses, an embedded device, and headphones for audio feedback. 
		}
	\label{321}
\end{figure} 

Based on visual sensors, we have designed a wearable steering assistance to guide people with impaired vision in independent movement within a representative sports scenario, the athletics track. The hardware of the assistance is illustrated in  Fig. \ref{321} and the overview is shown in Fig. \ref{11}. 
The perception and navigation capabilities developed within the athletics track hold the potential for broad applicability in sports, fitness, and competitions.
However, they also come with challenges.
Detecting the track lines presents difficulties because they are typically thin and can be easily obstructed by people ahead. Furthermore, there is an implicit correlation between the track lines and people obstacles. Additionally, the uniqueness of the scenario makes existing datasets cannot be directly used. The difference between curve and straight also poses a challenge for planning. 

To tackle these issues, we meticulously annotated an athletics track dataset for multitasks.
Meanwhile, we proposed a lightweight and efficient multitask network, enabling continuous track segmentation even in partial occlusion and the detection of people obstacles, bridging the gap between the two tasks. 
Moreover, we employ a combination of sampling and spline curves to effectively address the problem of curves. Our planner leverages the positions of track lines and obstacles as constraints to mitigate computational complexity. 
Even been deployed on embedded devices, our system operates efficiently, delivering swift and accurate directional instructions. Our main contributions are:
\begin{figure*}[t]
	\centerline{\includegraphics[width=5.88in]{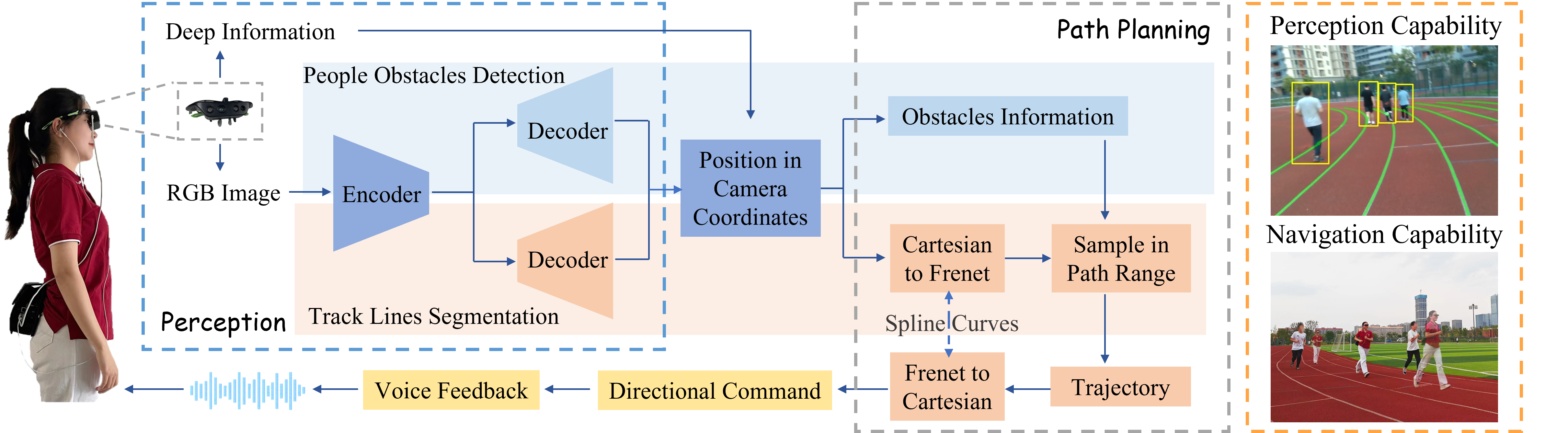}}
	\caption{The overview of the wearable steering assistance. The perception module utilizes a multitask network to extract track line and people positions from camera input, which are then used by the path planning module to generate a planned trajectory and provide it as voice feedback.}
	\label{11}
\end{figure*}
\begin{itemize}
	\item We designed a lightweight and efficient multitask network to achieve efficient and accurate obstacle detection and track line segmentation in embedded device. Additionally, we curated and annotated a dataset tailored for multitask in athletics track.
	\item We developed a wearable steering assistance for people with impaired vision in jogging, enabling users achieving free movement of 400-meter at an average speed of 1.34 m/s, meeting the level of normal people.
\end{itemize}
 
\section{RELATED WORKS}
Based on visual sensors, many studies have developed devices suitable for people with impaired vision.
\cite{23} uses visual and tactile feedback to assist the BVI in tasks of navigating mazes, locating chairs, and moving through crowds.
\cite{re1} use guidance robot to provide hospital navigation for people with impaired Vision.
\cite{ic20} uses a visual positioning system to assist in indoor navigation for the BVI.
\cite{21} employs an RGB-D camera to detect blind lanes and obstacles, using an improved artificial potential field method for navigation.
\cite{12} can meet the needs of navigation, target localization, facial recognition, and text reading, attaining speeds of 0.21 m/s during navigation tasks.
\cite{sr} adds a camera to the white cane for object positioning, and combines multiple sensors to assist people with impaired vision in improving their walking speed of 0.6 m/s.
Relevant researches have greatly facilitated the lives of the BVI and improved their mobility.
But they cannot meet the demand for sports.
Assisting people with impaired vision in achieving higher-speed mobility within the athletics track needs to accomplish two essential tasks: environmental perception and path planning.

Detection and segmentation methods\cite{wbl1}\cite{wbl2} can be employed for the perception of obstacles captured by RGB-D camera.
In terms of track line detection, we draw inspiration from lane detection in autonomous driving.
Deep learning-based lane detection methods include detection-based methods like LaneATT \cite{25} and Ultra-Fast-Lane-Detection\cite{26}. Segmentation-based methods include SCNN \cite{27} and RESA\cite{28}. While segmentation-based methods may struggle with occlusion issues, the detected lane lines are accurate for positioning. YOLOP\cite{212} proposed using a single network to simultaneously perform tasks like traffic object detection, drivable area segmentation, and lane detection. YOLOPv2\cite{v2} further improves the performance and speed of each task. In terms of training data, there are many publicly available datasets in autonomous driving, such as BDD100K \cite{411}, Cityscapes \cite{city}. However, there is no dataset specifically designed for athletics track.

In terms of path planning, purely relying on a single method has its limitations, many research began to integrate different planning methods to achieve better results.
Baidu Apollo's EM planner \cite{220} employs optimization function combined with polynomial curve for trajectory planning, utilizing dynamic programming and quadratic programming for path and speed planning respectively.\cite{222} combines RRT and Dijkstra in path planning, where RRT is employed to reduce the planning area, while Dijkstra's optimizer is used to obtain the final output path. \cite{ra20} combines graph-based search methods with sampling-based methods to quickly obtain the optimal solution.

\section{methodology}

\subsection{System Overview} 
 
The overview of the assistance is shown in Fig. \ref{11}. 
The user is equipped with an RGB-D camera, which continuously captures real-time color and depth images. 
These inputs are processed within the perception module, which employs a multitask network with two decoder heads to perform track lines segmentation and obstacles detection, effectively establishing connections between the track and people.
In path planning module, we employ spline curves to facilitate the transformation between Frenet and Cartesian coordinates, addressing curve challenges.
Within Frenet coordinates, we use the track boundaries and obstacle positions as constraints for average sampling, reducing the sampling area and lowering the computational burden. We employ dynamic programming to calculate the minimum path cost between sampling points, yielding a locally optimal trajectory.
Finally, the path planning results are transformed into directional commands, guiding the user through voice feedback.

\subsection{Hardware Setup}
 
As shown in Fig. \ref{321}, the wearable steering assistance comprises a pair of RGB-D glasses, an embedded device, and audio feedback. 
The wearable glasses are equipped with a RealSense D435i depth camera encased in a 3D-printed housing, enabling real-time capture of both RGB and depth images.
The embedded device is Jetson Orin NX, equipped with a processing capability of 100 TOPS.
This device is responsible for processing camera inputs and converting them into voice output.
The system is powered by a lithium battery and can operate continuously for 5 hours, meeting the endurance requirements for outdoor activities. 
The glasses themselves weigh 125g, while the embedded device weighs 274g and the overall weight is 1 kg, making it convenient to carry. 
The software is deployed on the Jetson Orin NX, eliminating the need for a remote server. The whole set of equipment fulfills the portability requirements.

\subsection{Dataset Construction}\label{dat}

\begin{figure}[htbp]
	\centerline{\includegraphics[width=3in]{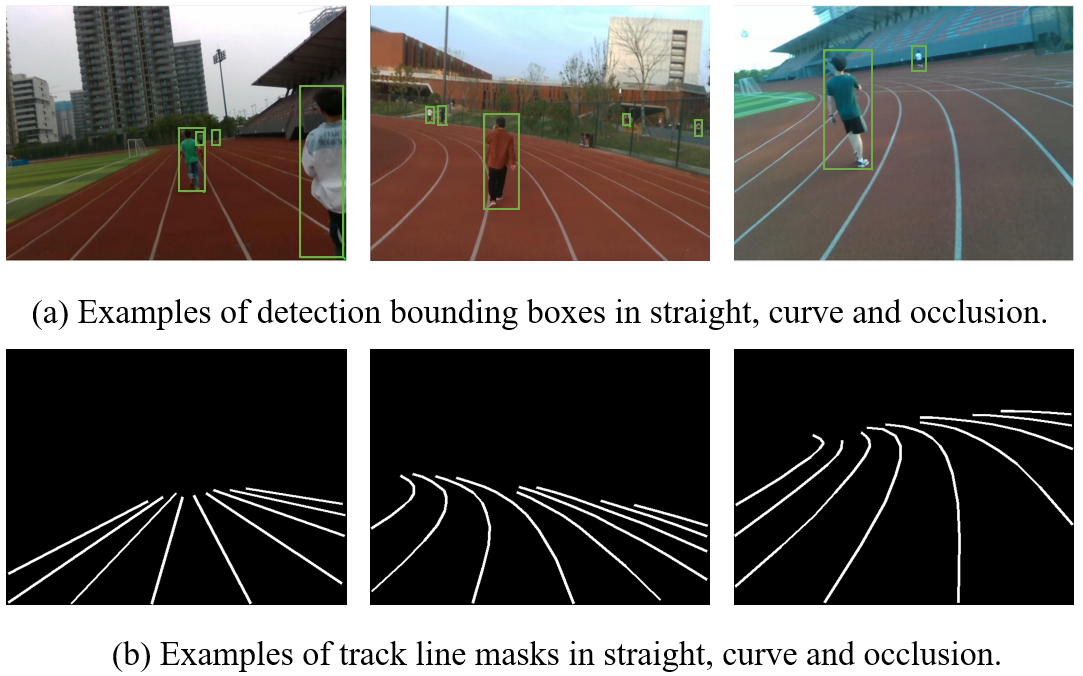}}
	\caption{Examples of dataset annotations.}
	\label{dataset}
\end{figure}

Given the uniqueness of the athletics track, we undertook the task of collecting, annotating, and curating an athletics track dataset, termed as Multi-data for Athletics Track (MAT), specifically tailored for multi-task detection. Our dataset comprises 1k track images, divided into training, evaluation, and test subsets at an 8:1:1 ratio. Each of these images was annotated with detection bounding boxes and track line masks. 
Some examples are shown in Fig. \ref{dataset}.

Our dataset features 2.47k instances of person. In terms of track line annotations, we have 376 images in straight and 624 images in curve. We performed continuous annotations on the track that were partial occlusion, resulting in 319 images with occlusions and 681 images without occlusion.
%

\subsection{Multi-task Network}\label{network}
\begin{figure}[t]
	\centerline{\includegraphics[width=3in]{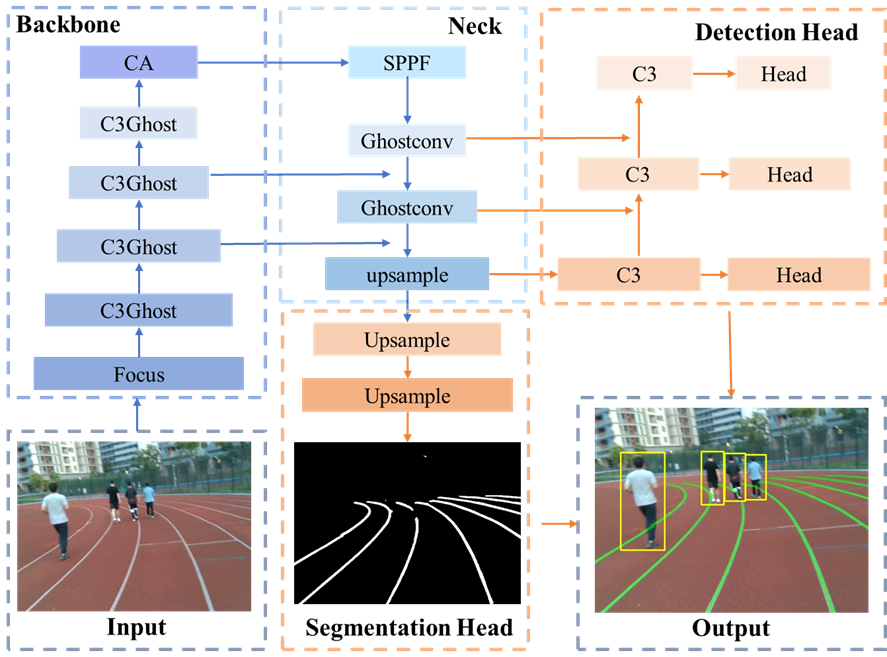}}
	\caption{The architecture of RunnerNet. It shares one encoder and combines two decoders to solve different tasks.}
	\label{331}
\end{figure}
We aim to achieve precise environment perception while operating on embedded devices. 
However, given the constraints of computational cost and memory, employing separate models for each task is impractical. 
To address this, we turn to multi-task networks, which leverage a shared backbone to extract valuable features and optimize performance across different tasks with separate decoders.
Therefore, we adopted an efficient and lightweight multi-task network capable of simultaneously handling track line segmentation and people detection tasks.
YOLOP \cite{212} is capable of simultaneously performing tasks like traffic object detection, drivable area segmentation, and lane detection. However, it cannot be directly applied in the athletics track due to varying task requirements and speed constraints. To alleviate the challenge, we made a series of improvements to it.

Firstly, our tasks are different. We focus on people detection and track line segmentation for planning. Therefore, in addition to the data labeling efforts mentioned in section \ref{dat}, we pruned down the detection head used for drivable area segmentation. 
The modified network now includes a shared encoder and two separate decoders for different tasks.
Secondly, we needed to meet the speed requirements for detection on embedded devices. Hence, we replaced the backbone with a lightweight architecture suitable for embedded devices, GhostNet \cite{332}. Additionally, we replaced the spatial pyramid pooling module with spatial pyramid pooling-fast to reduce the computational load while maintaining performance. 
The above design may lead to a decrease in feature extraction capability. 
Consequently, we introduced the Coordinate Attention(CA) \cite{333} at the end of the backbone to balance speed and accuracy without increasing computational demands.

In the detection head, we utilize FPN to propagate semantic features top-down and PAN for localization feature propagation bottom-up, achieving effective fusion.
In the track line segmentation head, we harnessed deep information from FPN. Through three upsampling processes, we acquire high-precision outputs, ensuring the continuity of track line predictions even in cases of partial occlusion.

we name our network RunnerNet, which substantially reduces the number of parameters and floating point operations compared to the original network. The network structure is shown in Fig. \ref{331}.

\subsection{Path Planning and Navigation}

Given the track's unique characteristics, we employ an approach to combine averaged sampling with spline curves, which effectively leverages track information.

This strategy employs spline curves to convert track lines from Cartesian coordinates to Frenet coordinates, facilitating path planning within the Frenet coordinates. 
We confine the sampling region within the track lines, limiting the planning space and reducing computational complexity. 
Uniform sampling is then performed within this track region, and sample points are selected by loss function to yield locally optimal trajectory points.
Finally, the trajectory is transformed back to Cartesian coordinates. The yaw angle of the path is converted into voice output, providing guidance to users. 

\subsubsection{Utilizing cubic spline for the transformation between Frenet and Cartesian} 
Representing curves in Cartesian coordinates is challenging. By transforming the curved boundaries from the Cartesian coordinates to the Frenet coordinates, nonlinear constraints can be converted into linear constraints. However, directly utilizing formulas for conversion is complex.  
We employ cubic spline curves $ {S_{\rm{i}}}(x) = {a_i} + {b_i}x + {c_i}{x^2} + {d_i}{x^3}$ for implicit solving. Unknown parameters $a_i$, $b_i$,  and $d_i$ can be represented by $c_i$.

Performing cubic spline separately on the $x$ and $y$ coordinates, the curve can be uniquely represented as:
\begin{equation}
	S(t) = \left( {x(t),y(t)} \right),
\end{equation}
where $S(t)$ represents the length of the curve in the Frenet coordinates, $x(t)$ and $y(t)$ is the cubic spline curves constructed under varying parameters. 
The arc length $S$ and yaw angle $\psi$ of the curve can be derived as:
\begin{equation}
	S = \sqrt {{{x'}^2} + {{y'}^2}},
\end{equation}
\begin{equation}
	\psi  = \arctan (\frac{{y'}}{{x'}}).
\end{equation}
where ${x'}$, ${y'}$ is the derivative of $x$, $y$.
Given the length $S_i$, along with the segment interval $i$ and interval parameters $a_i$, $b_i$, $c_i$, and $d_i$, it is possible to reverse the transformation from Frenet coordinates to Cartesian coordinates to obtain the  $x$ and $y$ coordinates. 

\subsubsection{Averaging sampling and cost function} 
EM planner\cite{220} defines a reference line to connect global planning with local planning.
What sets us apart is that the detected track lines provide natural constraints for sampling. Utilizing spline curves, we represent the two boundaries of the track in the Frenet coordinates and incorporate the positions of obstacles. With knowledge of the track width and forward viewing distance, we uniformly sample $n \times m$ points both laterally and longitudinally.
While traditional average sampling requires large computation, the constraints of tracks significantly reduce the computational workload. 
Unlike road environments, average sampling and calculating cost function within the narrow space of the track can provide more comprehensive information.

After constructing the sampling space, each point could be evaluated by the summation of the cost function. The cost $C_{total}$ between sample points can be expressed as:
\begin{equation}
	{C_{total}} = {C_{dis}} + {C_{obs}},
\end{equation}
where $C_{dis}$ represents the distance cost between the current sampling point ${s_1}$ and the next sampling point ${s_2}$, and $C_{dis}$ can be defined as:
\begin{equation}
	{C_{dis}} = \left\| {\left. {{s_1} - {s_2}} \right\|} \right..
\end{equation}
When there are obstacles, beyond the safe distance $d_{safe}$, the collision cost $C_{obs}$ is inversely proportional to the collision distance $d_{col}$, the scale factor is $k$, $C_{obs}$ can be defined as:
\begin{equation}
{C_{obs}} = \left\{ {\begin{array}{*{20}{c}}
		0&{{\rm{no obstacle}}}\\
		{\frac{{{k}}}{{{d_{col}}}}}&{{d_{col}} > {d_{safe}}}\\
		\infty &{{d_{col}} < {d_{safe}}}
\end{array}} \right.	
\end{equation}

We define the cost of the sampling points in the last row as the distance from the middle point.
Using the dynamic programming, we iterate the cost between sampling points from the last row forward. We set the initial point as the origin of the camera coordinates and iterate from front to back, selecting the sampling points with the minimum cost for each row as the trajectory points.
Finally, we use the cubic spline curve to fit the trajectory points to form a smooth path and obtain the yaw angle of the planned path by coordinate transformation.

\subsubsection{Audio Feedback} 
Effective feedback to the BVI is the key to achieving human-computer interaction and ensuring the availability of the system. We divided the visual field in front of the BVI into five sections corresponding to five direction instructions: forward, left forward, right forward, turn left, and turn right. The yaw angle of the planned output will fall into the corresponding interval, and the planning result will be converted into the direction instruction through voice feedback. In order to ensure real-time feedback, the trajectory planning is open-loop, and one direction instruction will be output for each frame.

\section{experiments}
In this section, we introduce the dataset setting, parameter setting, and experimental content. Experiments in section \ref{data} and section \ref{net} were carried out based on the configuration environment of NVIDIA GeForce RTX3060.
We ran all tests under the same experimental settings and evaluation metrics.
In section \ref{out}, we conducted outdoor experiments on the embedded device, Jetson Orin NX. 

\subsection{Settings}\label{data}
\begin{figure}[htbp]    
	\centering           
	\subfloat[Trained on BDD100K]  
	{	\label{411(a)}\includegraphics[width=0.19\textwidth]{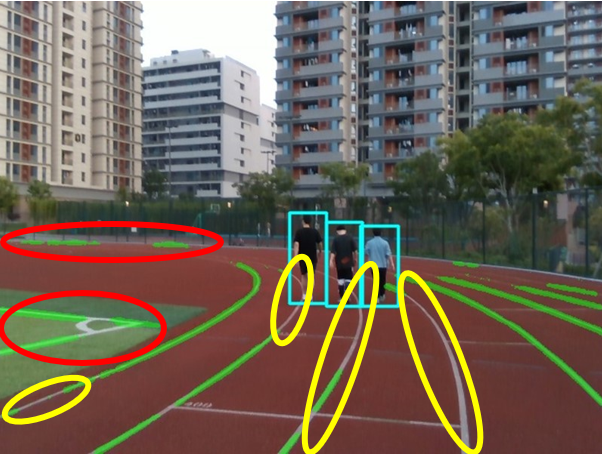}	}
	\subfloat[Trained on MAT]
	{	\label{411(b)}\includegraphics[width=0.19\textwidth]{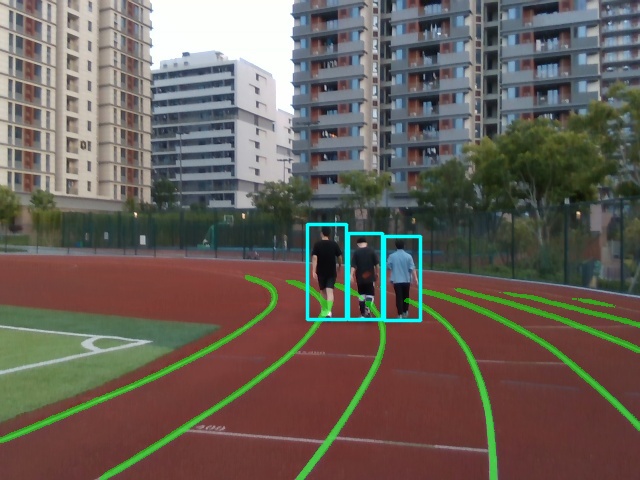}	}
	\caption{The valuation result trained on different dataset. The yellow ellipses are the false nagative and the red ellipses indicate the false positive.}   
	\label{411}            
\end{figure}

The BDD100K dataset \cite{411} is used for multi-task detection in autonomous driving. Although the algorithms trained on the BDD100K dataset have strong transferability,
as shown in Fig. \ref{411}, due to the uniqueness of the athletics track, the model trained on BDD100K cannot be directly used for track.

We compared the detection performance of models trained on BDD100K, on our dataset MAT, and further fine-tuned on MAT after BDD100K's pre-traind. The results are presented in Table \ref{tb411}.
Utilizing BDD100K as a pre-trained model led to improvements in track line segmentation and yielded people detection results similar to models trained on our dataset. Consequently, we evaluate network performance by using models pre-trained on BDD100K and fine-tuned on MAT.

We resize images in BDD100K from $1280\times720$ to $640\times384$, and the images in MAT is $640\times480$.
In pre-training, we employ an Adam optimizer for 240 epochs with a 3-epoch warm-up, a batch size of 24, an initial learning rate of 0.001, and a weight decay of 0.0005. For fine-tuning, we train the models for an additional 240 epochs with a batch size of 24, and a learning rate of 0.001.

\begin{table}[htbp]
	\centering
	\caption{Detection performance in different datasets}
	\begin{center}
		\begin{tabular}{c|cccc}
			\toprule
			Datasets     & Acc(\%) & mIoU(\%)  & Recall(\%)  & mAP$_{50}$(\%)  \\
			\midrule
			BDD100K      & 55.7    &68.0     & 56.6    & 51.7  \\
			MAT         & 87.1    &  82.5    &  87.7   & \textbf{84.7}  \\
			BDD100K+MAT  & \textbf{90.1}    & \textbf{82.9}     &  \textbf{87.7}  &  84.2  \\
			
			\bottomrule
		\end{tabular}
		\begin{tablenotes}
			\item The valuation is based on the model ($-$Head).
		\end{tablenotes}
		\label{tb411}
	\end{center}
\end{table}

\subsection{Performance of Multitask Networks}\label{net}
\begin{figure}[t]
	\centering
	\subfloat[Continuous segment]{
		\begin{minipage}[t]{0.3\linewidth}
			\centering
			\includegraphics[width=1.08in]{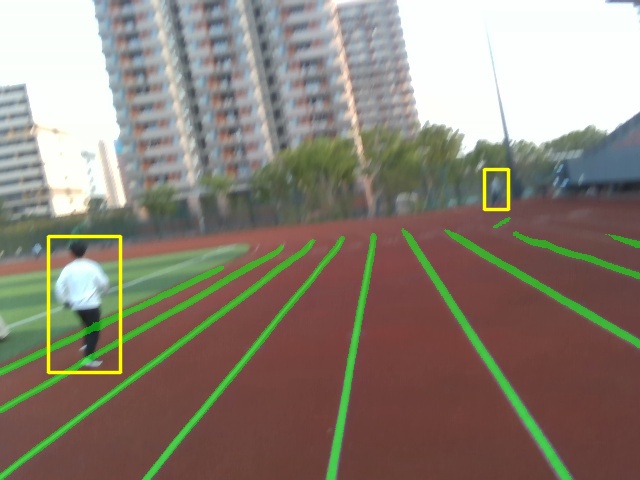}\\
			\vspace{0.05cm}
			\includegraphics[width=1.08in]{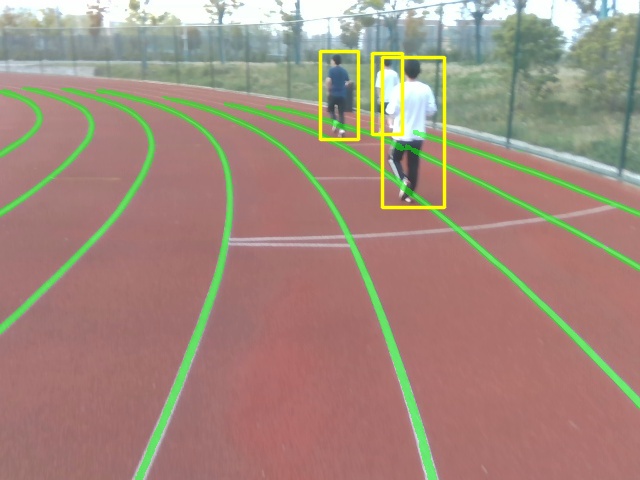}\\
			\vspace{0.02cm}
		\end{minipage}%
		\label{421c}	}%
	\subfloat[Multi-detection]{
		\begin{minipage}[t]{0.3\linewidth}
			\centering
			\includegraphics[width=1.08in]{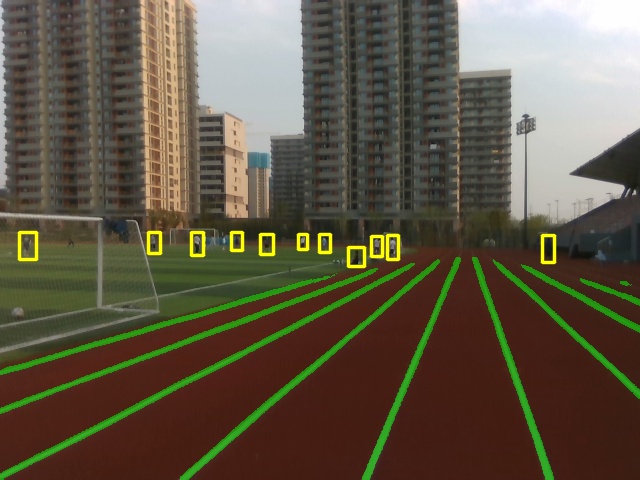}\\
			\vspace{0.05cm}
			\includegraphics[width=1.08in]{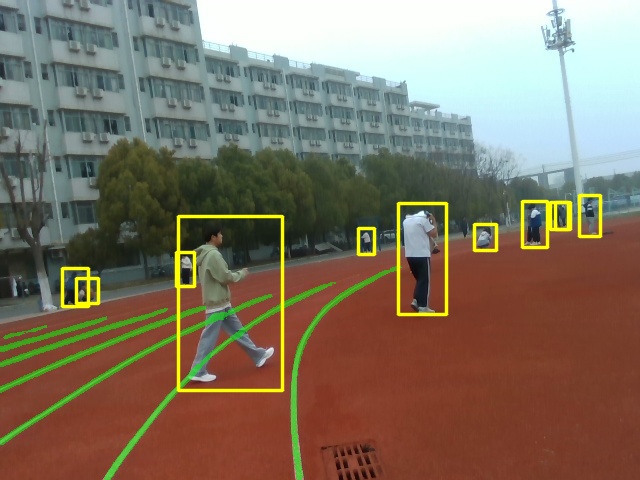}\\
			\vspace{0.02cm}
		\end{minipage}%
		\label{421d}	}%
	\subfloat[Night time]{
		\begin{minipage}[t]{0.3\linewidth}
			\centering
			\includegraphics[width=1.08in]{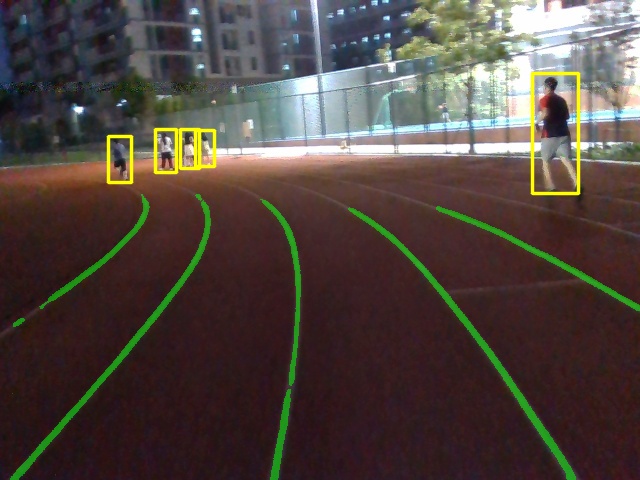}\\
			\vspace{0.05cm}
			\includegraphics[width=1.08in]{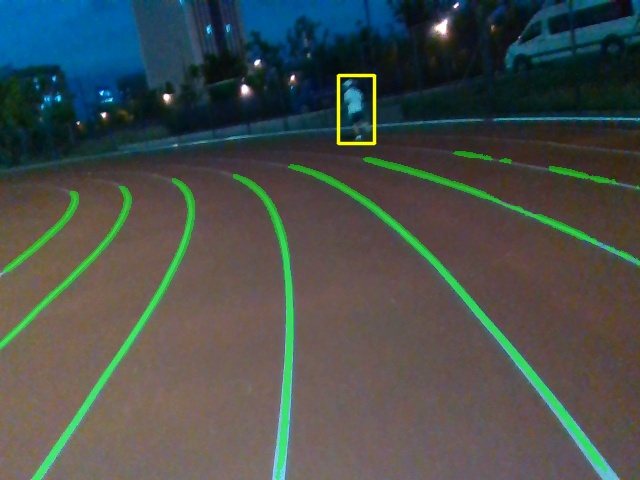}\\
			\vspace{0.02cm}
		\end{minipage}%
		\label{421f}	}%
	\centering
	\caption{Visualization of line segmentation and people detection results in multiple situations.
	}
	\vspace{-0.2cm}
	\label{421}
\end{figure}

The visualization of the network detection result is illustrated in  Fig. \ref{421}.
In section \ref{network}, we made improvements by pruning detection heads ($-$Head), changing the backbone ($+$GhostNet), and introducing attention mechanisms (RunnerNet). 
The impact of parameters and the performance of networks are shown in Table \ref{tb421}. 
We use Acc to evaluate the segmentation performance and use Recall and mAP$_{50}$ as the evaluation metric for detection accuracy.
Since the lack of prior works for multitask detection in the athletics track, we compared the performance of modified networks ($-$one head, $+$GhostNet, and RunnerNet).
The results indicate that our network reduced the FLOPs of the original network by $44\%$ and improved the speed by $35\%$.
Although the modified model is smaller than the original network, it has achieved higher detection accuracy. 
RunnerNet achieved $92.6\%$ accuracy in track line segmentation and achieved better detection performance compared to model ($-$Head). 
But its performance in detection is slightly lower than the model ($+$GhostNet), possibly due to its focus on the runway line.
To ensure the accuracy of track segmentation for the planning module, we choose RunnerNet, which performs best in track line segmentation, as the final model.

\begin{table}[htbp]
	\centering
	\caption{Ablation studies of network detection results}
	\begin{center}
		\begin{tabular}{c|cccccccc}
			\toprule
			Network &  FLOPs(G) &FPS & Acc(\%) &  Recall(\%)  & mAP$_{50}$(\%) \\
			\midrule
			YOLOP       & 15.64 & 54 & -  & -  & - \\
			$-$Head    & 12.05 & 69 & 90.1    &  87.7  &  84.2   \\
			$+$GhostNet       & 8.70 & 73 &   91.2  &    \textbf{88.5}  & \textbf{86.0}  \\
			RunnerNet & \textbf{8.70} & \textbf{73} & \textbf{92.6}    &  88.1   &  85.6 \\
			\bottomrule
		\end{tabular}
		
		\label{tb421}
	\end{center}
\end{table}

\subsection{Outdoor Experiments}\label{out}

To test the real-world usability of the system, three healthy subjects (User1, female, User2, male, and User3, male) were invited to wear the assistance for experiments, all participants wore blindfolds for the same level of vision.

In path planning, the sampling distance in the forward direction is 10 meters in Cartesian coordinates, with a sampling interval of 1 meter in Frenet coordinates and 5 sampling points in the horizontal direction. 

\begin{figure}[htbp]
	\centerline{\includegraphics[width=3.4in]{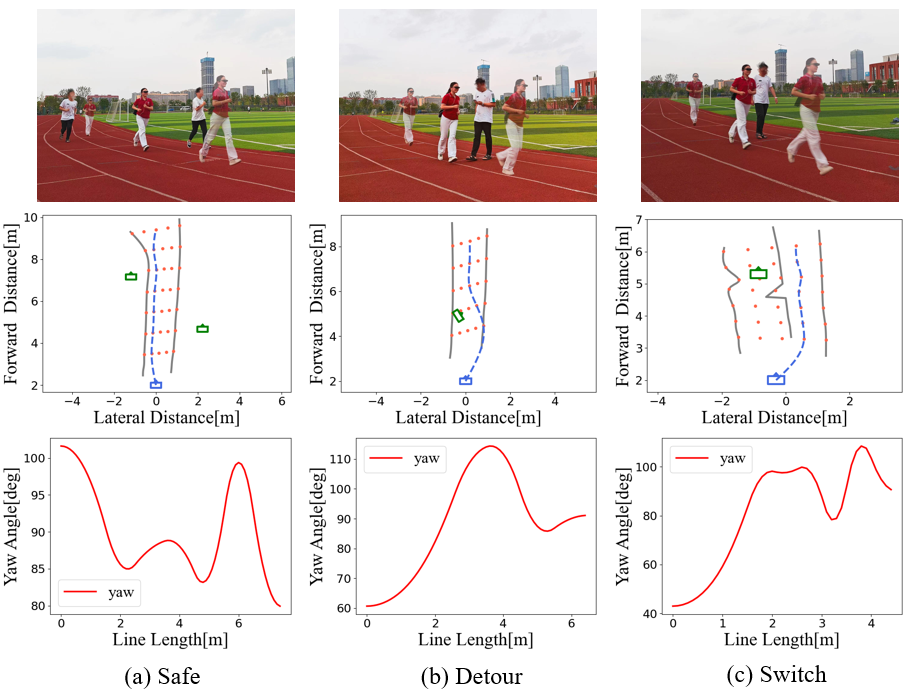}}
	\caption{Path planning experiments results in three sports scenarios. In each scenario, the first row represents the continuous trajectory changes of the user. The second row represents the path planning  results in real environment. The last row shows the changes of the yaw angle of the planned trajectory, providing directional instructions to the user.}
	\label{exp}
\end{figure}

We categorized the sports scenarios into three situations:
\begin{itemize}
	\item \textit{Safe}: no obstacle present, or obstacles have no impact.
	\item \textit{Detour}: detour around the obstacle and return.
	\item \textit{Switch}: change different track to bypass obstacles.
\end{itemize}
As shown in Fig. \ref{exp}, we conducted tests to evaluate the users' performance in different sports scenarios.
Users were able to complete tasks of moving forward, detouring, and changing tracks with assistance.

In addition to testing their performance in various sports scenarios, subjects also wore the assistance for free movement experiments, including 100-meter straight, 100-meter curve, and  400-meter track. The number of experiments and average speeds are shown in Table \ref{epx}.
\begin{table}[htbp]
	\centering
	\caption{Users performance in experiments}
	\begin{threeparttable}
			\setlength{\tabcolsep}{3.5mm}{
				\begin{tabular}{l|cccc}
					\toprule
					Users & User1 &User2   &User3 &   Average \\
					\midrule
					\textbf{100m straight} &        &	         &        & \\
					Num of exps &        6      & 3	         & 5       & 5\\
					average speed(m/s)  & 1.70	&1.50	& 1.22	& 1.52 \\  
					\midrule 
					\textbf{100m curve} &        &	         &        & \\
					Num of exps &        5      &4	         & 5       & 5\\
					average speed(m/s)      & 1.06     	&1.60	& 1.34 	& 1.23 \\  
					\midrule 
					\textbf{400m track} &        &	         &        & \\
					Num of exps &        3     & 2 	      & 3       & 3 \\
					average speed(m/s)  & 1.34	     	&1.50	&  1.23 & 1.34\\  
					\bottomrule
			\end{tabular}}
			
	\end{threeparttable}
	\label{epx}
\end{table}
The results indicate that, with the assistance of the device, users can achieve an average movement speed of 1.52 m/s in straight, 1.23 m/s in curve, and 1.34 m/s in track, reaching the level of normal people in jogging. Due to the processing speed limitations of embedded devices, we may not reach the speed of quick running, but it is sufficient to provide people with impaired vision with opportunities to participate in sports.

\section{CONCLUSIONS}

Based on visual sensors, we have developed a wearable steering assistance designed for people with impaired vision in jogging. 
We conducted outdoor experiments on the athletics track, and the results show that our wearable assistance enables the people with impaired vision to achieve an average speed of 1.34 m/s in track. The increase in speed is attributed to the accuracy of perception and the operational speed.
While our device provides faster mobility, it may not meet the speed requirements of quick running. In the future, we will continue to enhance the sports experience for people with impaired vision.

\bibliographystyle{IEEEtran}
\bibliography{ref}

\begin{thebibliography}{10}
\providecommand{\url}[1]{#1}
\csname url@samestyle\endcsname
\providecommand{\newblock}{\relax}
\providecommand{\bibinfo}[2]{#2}
\providecommand{\BIBentrySTDinterwordspacing}{\spaceskip=0pt\relax}
\providecommand{\BIBentryALTinterwordstretchfactor}{4}
\providecommand{\BIBentryALTinterwordspacing}{\spaceskip=\fontdimen2\font plus
\BIBentryALTinterwordstretchfactor\fontdimen3\font minus
  \fontdimen4\font\relax}
\providecommand{\BIBforeignlanguage}[2]{{%
\expandafter\ifx\csname l@#1\endcsname\relax
\typeout{** WARNING: IEEEtran.bst: No hyphenation pattern has been}%
\typeout{** loaded for the language `#1'. Using the pattern for}%
\typeout{** the default language instead.}%
\else
\language=\csname l@#1\endcsname
\fi
#2}}
\providecommand{\BIBdecl}{\relax}
\BIBdecl

\bibitem{11}
J.~D. Steinmetz, R.~R. Bourne, P.~S. Briant, S.~R. Flaxman, H.~R. Taylor, J.~B.
  Jonas, A.~A. Abdoli, W.~A. Abrha, A.~Abualhasan, E.~G. Abu-Gharbieh
  \emph{et~al.}, ``Causes of blindness and vision impairment in 2020 and trends
  over 30 years, and prevalence of avoidable blindness in relation to vision
  2020: the right to sight: an analysis for the global burden of disease
  study,'' \emph{The Lancet Global Health}, vol.~9, no.~2, pp. e144--e160,
  2021.

\bibitem{14}
W.~Zou, G.~Hua, Y.~Zhuang, and S.~Tian, ``Real-time passable area segmentation
  with consumer rgb-d cameras for the visually impaired,'' \emph{IEEE
  Transactions on Instrumentation and Measurement}, 2023.

\bibitem{12}
G.~Li, J.~Xu, Z.~Li, C.~Chen, and Z.~Kan, ``Sensing and navigation of wearable
  assistance cognitive systems for the visually impaired,'' \emph{IEEE
  Transactions on Cognitive and Developmental Systems}, vol.~15, no.~1, pp.
  122--133, 2022.

\bibitem{sr}
P.~Slade, A.~Tambe, and M.~J. Kochenderfer, ``Multimodal sensing and intuitive
  steering assistance improve navigation and mobility for people with impaired
  vision,'' \emph{Science robotics}, vol.~6, no.~59, p. eabg6594, 2021.

\bibitem{sp}
J.~Guerreiro, D.~Sato, S.~Asakawa, H.~Dong, K.~M. Kitani, and C.~Asakawa,
  ``Cabot: Designing and evaluating an autonomous navigation robot for blind
  people,'' in \emph{Proceedings of the 21st International ACM SIGACCESS
  Conference on Computers and Accessibility}, 2019, pp. 68--82.

\bibitem{23}
H.-C. Wang, R.~K. Katzschmann, S.~Teng, B.~Araki, L.~Giarr{\'e}, and D.~Rus,
  ``Enabling independent navigation for visually impaired people through a
  wearable vision-based feedback system,'' in \emph{2017 IEEE international
  conference on robotics and automation (ICRA)}.\hskip 1em plus 0.5em minus
  0.4em\relax IEEE, 2017, pp. 6533--6540.

\bibitem{re1}
K.~Tobita, K.~Sagayama, and H.~Ogawa, ``Examination of a guidance robot for
  visually impaired people,'' \emph{Journal of Robotics and Mechatronics},
  vol.~29, no.~4, pp. 720--727, 2017.

\bibitem{ic20}
H.~Zhang and C.~Ye, ``A visual positioning system for indoor blind
  navigation,'' in \emph{2020 IEEE international conference on robotics and
  automation (ICRA)}.\hskip 1em plus 0.5em minus 0.4em\relax IEEE, 2020, pp.
  9079--9085.

\bibitem{21}
Z.~Han, J.~Gu, and Y.~Feng, ``Blind lane detection and following for assistive
  navigation of vision impaired people,'' in \emph{2023 International
  Conference on Advanced Robotics and Mechatronics (ICARM)}.\hskip 1em plus
  0.5em minus 0.4em\relax IEEE, 2023, pp. 721--726.

\bibitem{wbl1}
B.~Wang, F.~Zhang, and Y.~Zhao, ``Lch: fast rgb-d salient object detection on
  cpu via lightweight convolutional network with hybrid knowledge
  distillation,'' \emph{The Visual Computer}, pp. 1--18, 2023.

\bibitem{wbl2}
F.~Zhang, H.~Liu, X.~Duan, B.~Wang, Q.~Cai, H.~Li, J.~Dong, and D.~Zhang,
  ``Dslsm: Dual-kernel-induced statistic level set model for image
  segmentation,'' \emph{Expert Systems with Applications}, vol. 242, p. 122772,
  2024.

\bibitem{25}
L.~Tabelini, R.~Berriel, T.~M. Paixao, C.~Badue, A.~F. De~Souza, and
  T.~Oliveira-Santos, ``Keep your eyes on the lane: Real-time attention-guided
  lane detection,'' in \emph{Proceedings of the IEEE/CVF Conference on Computer
  Vision and Pattern Recognition}, 2021, pp. 294--302.

\bibitem{26}
Z.~Qin, H.~Wang, and X.~Li, ``Ultra fast structure-aware deep lane detection,''
  in \emph{Computer Vision--ECCV 2020: 16th European Conference, Glasgow, UK,
  August 23--28, 2020, Proceedings, Part XXIV 16}.\hskip 1em plus 0.5em minus
  0.4em\relax Springer, 2020, pp. 276--291.

\bibitem{27}
A.~Parashar, M.~Rhu, A.~Mukkara, A.~Puglielli, R.~Venkatesan, B.~Khailany,
  J.~Emer, S.~W. Keckler, and W.~J. Dally, ``Scnn: An accelerator for
  compressed-sparse convolutional neural networks,'' \emph{ACM SIGARCH computer
  architecture news}, vol.~45, no.~2, pp. 27--40, 2017.

\bibitem{28}
T.~Zheng, H.~Fang, Y.~Zhang, W.~Tang, Z.~Yang, H.~Liu, and D.~Cai, ``Resa:
  Recurrent feature-shift aggregator for lane detection,'' in \emph{Proceedings
  of the AAAI Conference on Artificial Intelligence}, vol.~35, no.~4, 2021, pp.
  3547--3554.

\bibitem{212}
D.~Wu, M.-W. Liao, W.-T. Zhang, X.-G. Wang, X.~Bai, W.-Q. Cheng, and W.-Y. Liu,
  ``Yolop: You only look once for panoptic driving perception,'' \emph{Machine
  Intelligence Research}, vol.~19, no.~6, pp. 550--562, 2022.

\bibitem{v2}
C.~Han, Q.~Zhao, S.~Zhang, Y.~Chen, Z.~Zhang, and J.~Yuan, ``Yolopv2: Better,
  faster, stronger for panoptic driving perception,'' \emph{arXiv preprint
  arXiv:2208.11434}, 2022.

\bibitem{411}
F.~Yu, H.~Chen, X.~Wang, W.~Xian, Y.~Chen, F.~Liu, V.~Madhavan, and T.~Darrell,
  ``Bdd100k: A diverse driving dataset for heterogeneous multitask learning,''
  in \emph{Proceedings of the IEEE/CVF Conference on Computer Vision and
  Pattern Recognition}, 2020, pp. 2636--2645.

\bibitem{city}
M.~Cordts, M.~Omran, S.~Ramos, T.~Rehfeld, M.~Enzweiler, R.~Benenson,
  U.~Franke, S.~Roth, and B.~Schiele, ``The cityscapes dataset for semantic
  urban scene understanding,'' in \emph{Proceedings of the IEEE conference on
  computer vision and pattern recognition}, 2016, pp. 3213--3223.

\bibitem{220}
H.~Fan, F.~Zhu, C.~Liu, L.~Zhang, L.~Zhuang, D.~Li, W.~Zhu, J.~Hu, H.~Li, and
  Q.~Kong, ``Baidu apollo em motion planner,'' \emph{arXiv preprint
  arXiv:1807.08048}, 2018.

\bibitem{222}
R.~Chen, J.~Hu, and W.~Xu, ``An rrt-dijkstra-based path planning strategy for
  autonomous vehicles,'' \emph{Applied Sciences}, vol.~12, no.~23, p. 11982,
  2022.

\bibitem{ra20}
M.~P. Strub and J.~D. Gammell, ``Advanced bit (abit): Sampling-based planning
  with advanced graph-search techniques,'' in \emph{2020 IEEE International
  Conference on Robotics and Automation (ICRA)}.\hskip 1em plus 0.5em minus
  0.4em\relax IEEE, 2020, pp. 130--136.

\bibitem{332}
K.~Han, Y.~Wang, Q.~Tian, J.~Guo, C.~Xu, and C.~Xu, ``Ghostnet: More features
  from cheap operations,'' in \emph{Proceedings of the IEEE/CVF Conference on
  Computer Vision and Pattern Recognition}, 2020, pp. 1580--1589.

\bibitem{333}
Q.~Hou, D.~Zhou, and J.~Feng, ``Coordinate attention for efficient mobile
  network design,'' in \emph{Proceedings of the IEEE/CVF Conference on Computer
  Vision and Pattern Recognition}, 2021, pp. 13\,713--13\,722.

\end{thebibliography}

\addtolength{\textheight}{-12cm}   

\end{document}